\documentclass{bmvc2k}

\usepackage{booktabs, multirow, makecell}
\usepackage{wrapfig}
\usepackage{amsmath}
\usepackage{amsfonts}
\usepackage{bbding}
\usepackage{etoolbox}

\definecolor{bmvc_blue}{RGB}{0, 0, 102}
\usepackage{caption}
\title{Boost Video Frame Interpolation via \\ Motion Adaptation}

\addauthor{Haoning Wu}{whn15698781666@sjtu.edu.cn}{1}
\addauthor{Xiaoyun Zhang$^\dagger$}{xiaoyun.zhang@sjtu.edu.cn}{1}
\addauthor{Weidi Xie}{weidi@sjtu.edu.cn}{1,2}
\addauthor{Ya Zhang}{ya_zhang@sjtu.edu.cn}{1,2}
\addauthor{Yanfeng Wang$^\dagger$}{wangyanfeng622@sjtu.edu.cn}{1,2}

\addinstitution{
 Coop. Medianet Innovation Center,\\
 Shanghai Jiao Tong University, China\\
}
\addinstitution{
 Shanghai AI Laboratory, China\\
}

\runninghead{WU, ET.AL}{Boost Video Frame Interpolation via Motion Adaptation}

\def\etal{\emph{et al}\bmvaOneDot}

\newcommand\blfootnote[1]{%
    \begingroup 
    \renewcommand\thefootnote{}\footnote{#1}%
    \addtocounter{footnote}{-1}%
    \endgroup 
}

\begin{document}

\maketitle
\blfootnote{$\dagger$: Corresponding author.}

\begin{abstract}

Video frame interpolation (VFI) is a challenging task that aims to generate intermediate frames between two consecutive frames in a video. 
Existing learning-based VFI methods have achieved great success, 
but they still suffer from limited generalization ability due to the limited motion distribution of training datasets. 
In this paper, we propose a novel optimization-based VFI method that can adapt to unseen motions at test time. Our method is based on a {\em cycle-consistency adaptation} strategy that leverages the motion characteristics among video frames. 
We also introduce a lightweight {\em adapter} that can be inserted into the motion estimation module of existing pre-trained VFI models to improve the efficiency of adaptation. Extensive experiments on various benchmarks demonstrate that our method can boost the performance of two-frame VFI models, 
outperforming the existing state-of-the-art methods, 
even those that use extra input frames.
Project page: \href{https://haoningwu3639.github.io/VFI_Adapter_Webpage/}{https://haoningwu3639.github.io/VFI\_Adapter\_Webpage/}


\end{abstract}

\begin{figure}[t]
\footnotesize
  \centering  
    \includegraphics[width=.99\textwidth]{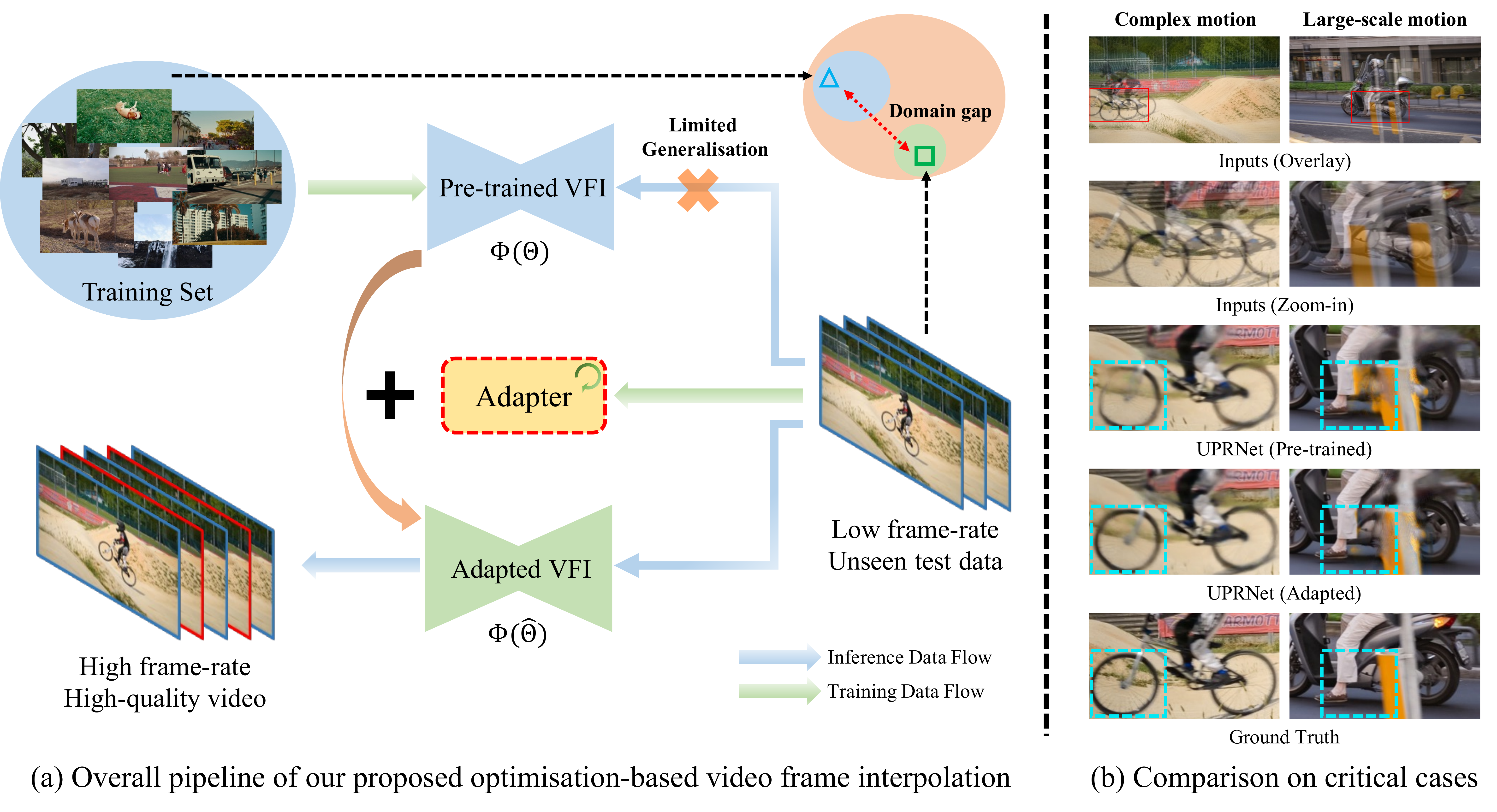}
     \vspace{-0.1cm}
  \caption{\textbf{High-level idea overview.} 
  (a) To address the generalisation challenge of VFI models due to domain gap on unseen data, we propose the optimisation-based video frame interpolation. By performing test-time motion adaptation on our proposed lightweight {\em adapter}, we achieve the generalization of VFI models across different video scenarios and subsequently boost their performance. 
  (b) Visual comparison on the cases with complex and large-scale motions from DAVIS~\cite{perazzi2016benchmark} dataset. Our method assists VFI models in generalising to diverse scenarios and synthesizing high-quality frames with clearer structures and fewer distortions.
  }
 \label{fig:teaser}
 \vspace{-0.4cm}
\end{figure}

\section{Introduction}
\label{sec:intro}

Video frame interpolation (VFI) is a technique that increases the temporal resolution of a video by synthesizing intermediate frames between existing frames. 
This results in smoother transitions between frames, which can improve the overall quality of the video. VFI has a wide range of applications, 
including video compression~\cite{wu2018video, peleg2019imnet}, 
slow-motion generation~\cite{liu2017dvf} and novel-view synthesis~\cite{flynn2016deepstereo, kalantari2016learning}, \textit{etc}.
In the literature, existing VFI approaches can be roughly divided into two categories: 
flow-agnostic and flow-based frame synthesis. 
Flow-agnostic approaches do not use optical flow to compensate for motion between frames, instead, they typically use a combination of adaptive convolution kernel and interpolation techniques. 
Flow-based approaches, on the other hand, do use optical flow to estimate the motion between frames. This information is then used to synthesize the intermediate frames. 
Recently, thanks to the rapid progress in optical flow estimation~\cite{sun2018pwc, teed2020raft}, flow-based VFI approaches~\cite{niklaus2018context, jiang2018super, bao2019memc, hu2022many, jin2022unified} have become the dominant approach in the field.

Existing flow-based frame interpolation models are learning-based, 
that usually follow a similar pipeline: extract visual features from the input frames, 
estimate optical flows between the reference and target frame to be synthesized, warp the input frames and their contextual features based on the estimated flows, then finally synthesize the intermediate frame based on the aligned visual features. 
Under such design, various model architectures have been explored, 
such as constructing multi-scale pyramids~\cite{kong2022ifrnet, huang2022real, jin2022unified} and designing vision transformer~\cite{shi2022vfit, lu2022vfiformer}. 
Additionally, the incorporation of extra input information, such as depth~\cite{bao2019depth}, or more adjacent frames~\cite{xu2019quadratic, liu2020enhanced, kalluri2023flavr, shi2022vfit}, have also been explored as potential solutions. 
Despite training such meticulously-designed models on a large amount of videos, 
generalisation towards real videos with complex and large-scale motions still remains challenging.

Unlike existing approaches on architecture design, 
we explore an alternative direction for boosting models' performance at inference time, 
{\em i.e.} optimisation-based frame interpolation via test-time motion adaptation. 
Test-time adaptation has been proven effective in enhancing the performance of models in specific scenarios, as demonstrated in numerous computer vision tasks, 
such as image super-resolution~\cite{michaeli2013nonparametric, shocher2018zero, huang2015single} and image deblurring~\cite{chi2021testdeblur}, 
{\em etc}, it remains unexplored in the domain of video frame interpolation.
To this end, we devise a novel strategy suitable for video frame interpolation, 
namely {\em cycle-consistency adaptation}. 
The key idea is to construct triplet samples with consecutive frames from low frame-rate videos during test time and optimise model parameters on each video sequence by leveraging the inter-frame consistency. Considering that the adaptation process necessitates considerable inference time, we further propose a simple, yet effective {\em adapter} that can be injected into the existing motion estimation module of VFI, enabling motion refinement with less than $4\%$ trainable parameters comparing to the original model. 

To summarise, we make the following contributions in this paper:
(i) to improve the generalisation ability of existing VFI models and boost their performance, we propose an optimisation-based motion adaptation strategy based on cycle-consistency, 
that allows to tune the model on each test video sequence at inference time;
(ii) to address the drawback of high time cost associated with test-time adaptation, 
we design a simple, yet effective plug-in {\em adapter} to refine the motion flow estimated by VFI models, with minimal tuning cost;
(iii) we experiment on various models and benchmarks, 
demonstrating that our optimisation-based method can always boost the performance of existing two-frame VFI models, even outperforming approaches with extra inputs.


\section{Related Work}



\vspace{3pt}
\noindent {\textbf{Video Frame Interpolation.}} 
Video frame interpolation (VFI) is a long-standing computer vision research topic and has been widely studied. 
The recent literature on training deep neural networks has demonstrated extraordinary performance for video frame interpolation. 
Depending on whether optical flow is used, 
VFI methods can be broadly classified into two categories: 
flow-agnostic~\cite{niklaus2017adaconv, niklaus2017sepconv, choi2020channel, cheng2021multiple, kalluri2023flavr, lee2020adacof, shi2021video} and flow-based ones~\cite{niklaus2018context, jiang2018super, niklaus2020softmax, park2020bmbc, park2021asymmetric, bao2019memc, hu2022many, jin2022unified, jin2023enhanced}.
With the rapid development of optical flow estimation algorithms~\cite{ilg2017flownet, hui2018liteflownet, sun2018pwc, teed2020raft}, flow-based approaches have taken the dominant position, 
which typically employ optical flow to warp visual features of adjacent frames to synthesize intermediate frame, hence the quality of the generated frame is highly affected by the accuracy of motion estimation.
Various strategies have been explored to improve the performance of flow-based methods. These include exploring depth information for occlusion reasoning~\cite{bao2019depth}, guiding the learning of motion estimation via knowledge distillation~\cite{huang2022real, kong2022ifrnet}, designing efficient architectures for high-resolution videos with relatively large motion~\cite{sim2021xvfi, reda2022film}, utilizing the long-range dependency modeling capability of transformer for processing extensive motion~\cite{lu2022vfiformer, shi2022vfit}, and making full use of multiple adjacent frames for complex motion modeling~\cite{xu2019quadratic, liu2020enhanced, kalluri2023flavr, shi2022vfit}. 

In contrast to the aforementioned learning-based methods that aim to enhance VFI model generalisation and performance by modifying model architecture or incorporating extra input information, we consider an optimisation-based video frame interpolation that adapts pre-trained VFI models to the motion patterns in different video sequences at inference time. 



\vspace{5pt}
\noindent {\textbf{Cycle Consistency.}} 
The idea of cycle consistency has been widely explored in various self-supervised methods, such as image representation and correspondence learning~\cite{wu2021contrastive, wang2019learning, lai2019selfsupervised, zhou2016learning, Dwibeditemporal}.
For low-level vision tasks, CycleGAN~\cite{zhu2017unpaired} utilizes cycle-consistency loss to constrain the training process of generative adversarial networks on image2image translation task. ARIS~\cite{zhou2022aris} exploits cycle-consistency constraint to augment the models' ability for arbitrary super-resolution. CyclicGen~\cite{liu2019cyclicgen} and Reda \etal \cite{reda2019unsupervised} learn video frame interpolation in an unsupervised manner with the proposed cycle-consistency loss on general video data. 

\vspace{5pt}
\noindent {\textbf{Test-time Adaptation.}} Test-time adaptation manages to adapt the trained model to test data distribution for performance improvement, and has been successful in tasks such as classification~\cite{sun2020test} and pose estimation~\cite{li2021test}, {\em etc}.
Moreover, the idea has also been widely employed in low-level vision, such as super-resolution~\cite{michaeli2013nonparametric, shocher2018zero, huang2015single}, in order to improve the generalisation ability of models on various data. For video frame interpolation, SAVFI~\cite{choi2020scene} utilizes a meta-learning framework for training models to achieve scene-adaptive frame synthesis.

In this paper, we exploit {\em cycle-consistency} for test-time motion adaptation in video frame interpolation task, optimising motion for test video sequences on-the-fly, thus improving performance steadily. Moreover, a lightweight yet effective plug-in {\em adapter} has been proposed to improve the efficiency of VFI test-time adaptation.




\section{Methods}

In this section, we start by introducing the considered problem scenario,
{\em e.g.}, learning-based and our proposed optimisation-based video frame interpolation (Sec.~\ref{sec3.1}). Subsequently, we detail our proposed motion adaptation strategy that is suitable for video frame interpolation, namely {\em cycle-consistency adaptation} (Sec.~\ref{sec:3.2}). Lastly, to further improve the efficiency of test-time adaptation, 
we present a lightweight {\em adapter} that can serve as a plug-in module for better motion estimation in VFI models (Sec.~\ref{sec:3.3}).

\subsection{Problem Scenario}
\label{sec3.1}
Given a low frame-rate input video, the goal of video frame interpolation is to synthesize intermediate frame between two or multiple adjacent frames, ending up of high-frame videos with smoother motion. 
In a learning-based VFI model, it typically takes two consecutive frames ($\mathcal{I}_{i-1}$ and $\mathcal{I}_{i+1}$) as input and outputs one single intermediate frame ($\mathcal{I}_{i}$) between them, the model's parameters are learnt by minimizing the empirical risk on a training set:
\begin{align}
\begin{split}
&\mathcal{L}_{\mathcal{D}}(\Theta) = \mathbb{E}_{\mathcal{D}} (||\hat{\mathcal{I}}_{i} - \mathcal{I}_i||) \qquad 
\text{where \hspace{0.06cm}} \hat{\mathcal{I}}_{i} = \Phi(\mathcal{I}_{i-1}, \mathcal{I}_{i+1}; \Theta)
\end{split}
\end{align}
$\Phi(\cdot)$ refers to the video frame interpolation model, 
$\Theta$ denotes the parameters to be learnt on a large-scale training set~($\mathcal{D}$), and $\{\hat{\mathcal{I}}_{i}\}$ denote the predicted intermediate frames. 
At inference time, the model is expected to generalise towards unseen videos. 
However, in practise, these models can sometimes be fragile on cases with diverse and complex motions.


In this paper, we consider to improve the model's efficacy, with optimisation-based video frame interpolation via test-time motion adaptation:
\begin{equation}
    \hat{\Theta}_{\mathcal{V}} = \arg \min_{\Theta}  \mathcal{L}_{\mathcal{V}}(\Phi, \mathcal{V}; \Theta)
\end{equation}
where $\Phi(\cdot)$ denotes a pre-trained VFI model with parameters $\Theta$, and we aim to further optimise its parameters to boost the performance on one given test video sequence~($\mathcal{V}$). 
In the following section, we aim to answer the core question:  
\textbf{how can we design the objective function~($\mathcal{L}_{\mathcal{V}}$), 
given only low frame-rate videos are presented at inference time ?} 

\subsection{Cycle-Consistency Adaptation}
\label{sec:3.2}

At inference time, we construct a series of triplet samples from the given test video sequence, each consists of three consecutive frames, 
for example, $\mathcal{D}_1 = \{\mathcal{I}_1, \mathcal{I}_3, \mathcal{I}_5\}$,
$\mathcal{D}_2 = \{\mathcal{I}_3, \mathcal{I}_5, \mathcal{I}_7\}$, {\em etc}.
Our goal is to optimise the model's parameters using $\mathcal{D}_1$, 
and to boost the performance on synthesizing intermediate frames $\{\hat{{\mathcal{I}}}_{2}, \hat{{\mathcal{I}}}_{4}\}$ by exploiting the cycle-consistency constraint.

\begin{figure}[t]
\footnotesize
  \centering  
    \includegraphics[width=\textwidth]{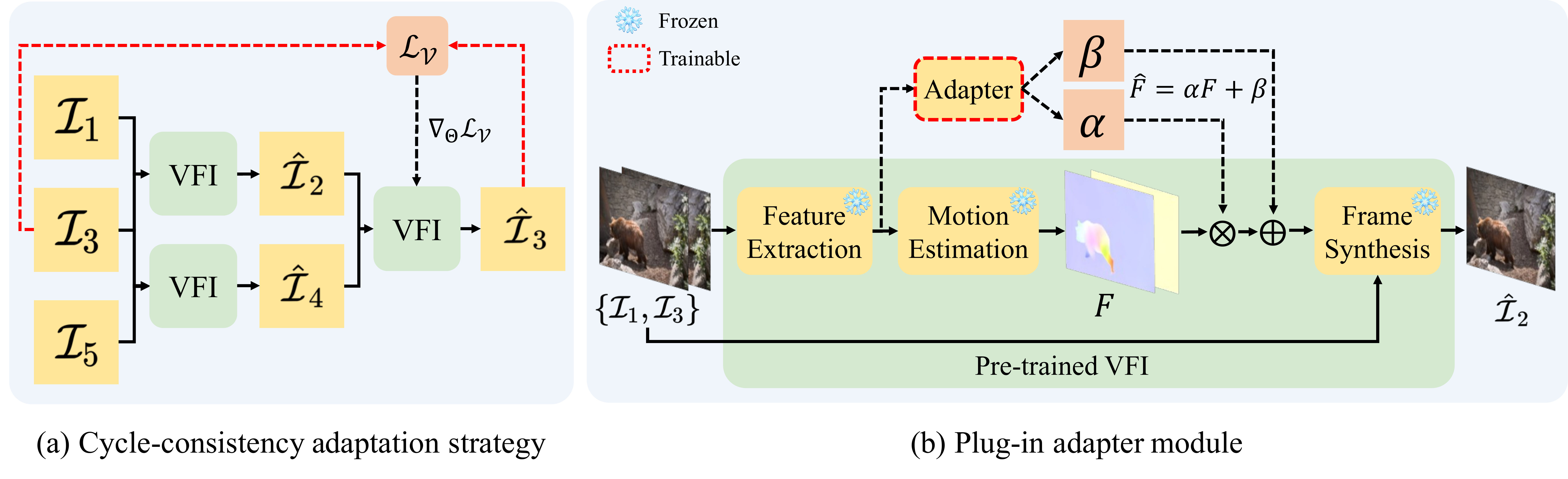}
\vspace{-0.4cm}
  \caption{\textbf{Proposed {\em cycle-consistency adaptation} strategy and plug-in {\em adapter} module for efficient test-time adaptation.} (a) {\em Cycle-consistency adaptation} first synthesizes intermediate frames between each two input frames and reuses them to interpolate the target frame to calculate cycle-loss, which fully utilizes the consistency within video sequences.
  (b) To improve efficiency, we freeze all the parameters of pre-trained VFI models and solely optimise the proposed plug-in {\em adapter}, which predicts a set of parameters $\{\alpha, \beta\}$ based on the extracted visual features. The pixel-wise weights $\alpha$ and biases $\beta$ are used for rectifying the estimated flow to fit each video sequence.}
 \label{fig:method}
 \vspace{-0.2cm}
\end{figure}

As detailed in Figure~\ref{fig:method} (a), 
taking the triplet $\mathcal{D}_1 = \{\mathcal{I}_1, \mathcal{I}_3, \mathcal{I}_5\}$ for demonstration, we can generate the intermediate frames with a pre-trained off-the-shelf VFI model on input consecutive video frames, 
\begin{equation}
    \begin{split}
         \hat{\mathcal{I}}_2 = \Phi(\mathcal{I}_1, \mathcal{I}_3; \Theta)  \qquad 
         \hat{\mathcal{I}}_4 = \Phi(\mathcal{I}_3, \mathcal{I}_5; \Theta)
    \end{split}
\end{equation}
and reuse the synthesized frames to predict the target frame: 
\begin{equation}
    \hat{\mathcal{I}}_3 = \Phi(\hat{\mathcal{I}}_2, \hat{\mathcal{I}}_4; \Theta)    
\end{equation}
 The model's parameters $\Theta$ can be updated  according to:
\begin{equation}
\begin{split}
    &\Theta \leftarrow \Theta - \eta \nabla_{\Theta} \mathcal{L}_{\mathcal{V}}(\hat{\mathcal{I}}_3, \mathcal{I}_3) \quad \text{where \hspace{0.04cm}} \mathcal{L}_{\mathcal{V}}(\Theta) = ||\hat{\mathcal{I}}_{3} - \mathcal{I}_3||
\end{split}
\end{equation}
$\eta$ denotes the adaptation learning rate. 
This cycle-consistency adaptation strategy enables the model to make full use of inter-frame consistency, thereby acquiring motion patterns that are more suitable for testing scenarios and achieving stable performance improvements. 

\subsection{Lightweight Motion Adaptation with Plug-in Adapter}
\label{sec:3.3}
Optimising the entire model for test-time motion adaptation incurs computation overhead, here, we further propose a simple, lightweight plug-in {\em adapter} module that can be inserted into the motion estimation module of existing pre-trained VFI models, requiring minimal tuning to boost the performance.
As depicted in Figure \ref{fig:method} (b), we freeze all parameters of the pre-trained VFI model, {\em e.g.}, feature extraction, motion estimation and frame synthesis modules, then incorporate the proposed adapter into the motion estimation module, which takes visual features extracted from adjacent frames as input, and predicts a set of parameters $\{\alpha, \beta\}$ to adjust motion estimation per video sequence. 


To be specific, the adapter reuses the convolutional features from motion estimation module, and subsequently employs a $1\times 1$ Convolution layer to transform the feature map into pixel-wise weights $\alpha$ and biases $\beta$:
\begin{equation}
    \hat{F} = \alpha F + \beta \text{,\hspace{0.1cm} where \hspace{0.04cm}}\{\alpha, \beta\} = \text{Conv}(\Psi(M))
\end{equation}
$M$ represents the extracted visual feature map and $\Psi$ denotes the resused convolutional layers, then the predicted $\{\alpha, \beta\}$ are used to modify the estimated motion flow $F$. During test-time adaptation, we only finetune the parameters in adapter, 
effectively refining the estimated motion flow and boosting the performance of video frame interpolation model.

\section{Experiments}

In this paper, we start from a series of VFI models that have been pre-trained on Vimeo90K-Triplet dataset~\cite{xue2019toflow}, then train our proposed {\em adapter} module on the same dataset.  
For test-time motion adaptation, the model is further optimised on each testing sequence from three benchmark datasets, including Vimeo90K-Septuplet~\cite{xue2019toflow}, DAVIS~\cite{perazzi2016benchmark}, and SNU-FILM~\cite{choi2020channel}. 

\vspace{5pt}
\noindent{\bf Training Set.} Vimeo90K-Triplet~\cite{xue2019toflow} training set comprises 51,312 triplets, wherein each triplet contains three consecutive video frames with a spatial resolution of $448 \times 256$ pixels.

\vspace{5pt}
\noindent{\bf Testing Set.} 
Vimeo90K-Septuplet~\cite{xue2019toflow} dataset encompasses 7,824 seven-frame sequences for testing, with a fixed spatial resolution of $448 \times 256$ pixels. 
DAVIS~\cite{perazzi2016benchmark} is a typically high-quality video segmentation dataset with a fixed resolution of $854 \times 480$ pixels. 
Following FLAVR~\cite{kalluri2023flavr} and VFIT~\cite{shi2022vfit}, 
we also report performance on 2,847 septuplet test samples from DAVIS. 
SNU-FILM~\cite{choi2020channel} dataset contains 1,240 triplets, 
with a predominant resolution of approximately $1280 \times 720$ pixels. 
It comprises of four categories with ascending motion scales: easy, medium, hard, and extreme. We further sample the surrounding frames of the ground truth and extend each triplet into a septuplet. 
As a result, the easy, medium, and hard categories contain 310 sequences,
while the extreme category contains 234 sequences. Among these septuplets in the above three benchmarks, the four odd frames compose the input video sequence of the VFI model, as described in Sec.~\ref{sec3.1}. And the intermediate frame is regarded as the ground truth for the frame to be interpolated in our experiments.

\vspace{5pt}
\noindent{\bf Evaluation Metrics.} 
Following common practise, we report Peak Signal-to-Noise Ratio (PSNR) and Structural-Similarity-Image-Metric (SSIM) on the RGB channel of the target interpolated frame for the three benchmark datasets.

\vspace{5pt}
\noindent{\bf Training Details. } 
To start with, we freeze the parameters of the pre-trained VFI models and only
train our proposed plug-in adapter on Vimeo90K for 30 epochs with a batch size of 16. We randomly crop $256 \times 256$ patches and augment the training data using horizontal and vertical flipping, temporal order reversing and RGB channel flipping. We use the AdamW~\cite{loshchilov2017fixing} optimizer with $\beta_1 = 0.9$ and $\beta_2 = 0.99$, and the learning rate is gradually reduced from $3 \times 10^{-4}$ to $3 \times 10^{-5}$ using cosine annealing during the whole training process. Considering the high resolution of some video data, optimisation-based VFI is performed on one 40G NVIDIA A100 GPU. During test-time adaptation for the full model, we use a fixed learning rate of $1 \times 10^{-5}$ and calculate L1 loss to fine-tune parameters of VFI models. As for adapter-boosted models, since other parameters have been frozen, we adapt the plug-in adapter module to each test sequence with a larger learning rate of $1 \times 10^{-4}$ for IFRNet~\cite{kong2022ifrnet}, $3 \times 10^{-4}$ for UPRNet~\cite{jin2022unified}, and $1 \times 10^{-3}$ for RIFE~\cite{huang2022real}, respectively.

\section{Results}

In this section, we start by providing experimental results for comparison with existing state-of-the-art approaches~(Sec.~\ref{sec:comp}), 
showing the effectiveness of our proposed {\em cycle-consistency adaptation}, in both end-to-end and plug-in adapter finetuning scenarios. 
After that, we conduct a series of ablation studies on the critical design choices on our adaptation strategy and the plug-in adapter module~(Sec.~\ref{sec:abl}).

\subsection{Comparison to state-of-the-art}
\label{sec:comp}

\noindent {\bf Quantitative Results.}  
We compare our boosted models with 9 representative learning-based models trained on Vimeo90K-Triplet~\cite{xue2019toflow}, including flow-free ones: SepConv~\cite{niklaus2017sepconv}, EDSC~\cite{cheng2021multiple} and FLAVR~\cite{kalluri2023flavr} and flow-based ones: RIFE~\cite{huang2022real}, UPRNet~\cite{jin2022unified} and \textit{etc}. Among them, FLAVR~\cite{kalluri2023flavr} and VFIT~\cite{shi2022vfit} take four frames as input, while others only use two adjacent frames. 
Specifically, we consider two scenarios, namely, end-to-end finetuning (e2e), 
or plug-in adapter finetuning (plugin), the former optimises all parameters in the model, 
denoted as [model-ours-e2e], 
while the latter only updates adapters, denoted as [model-ours-plugin]. 
By default, all test-time motion adaptations are only conducted for 10-step updates, 
with one exception on [model-ours-e2e++], which has performed 30-step adaptation, aiming to show the performance variation with more optimisation steps.

\begin{table}[t]
\scriptsize
\begin{center}
\tabcolsep=0.06cm
\begin{tabular}{lcccccccc}
\toprule
\multicolumn{1}{c}{\multirow{2}{*}{Methods}} & \multicolumn{2}{c}{Adaptation}  & \multirow{2}{*}{Vimeo90K~\cite{xue2019toflow}} & \multirow{2}{*}{DAVIS~\cite{perazzi2016benchmark}} & \multicolumn{4}{c}{SNU-FILM~\cite{choi2020channel}}                                                    \\ 
\cmidrule{2-3} \cmidrule{6-9} 
\multicolumn{1}{c}{}                         &  e2e    &  plugin                       &                           &                        & Easy                  & Medium                & Hard                  & Extreme \\ 
\midrule
SepConv~\cite{niklaus2017sepconv}      & \multicolumn{1}{c}{\XSolidBrush} & \multicolumn{1}{c|}{\XSolidBrush}     & \multicolumn{1}{c|}{33.72 / 0.9639}     & \multicolumn{1}{c|}{26.65 / 0.8611}  & \multicolumn{1}{c|}{40.21 / 0.9909} & \multicolumn{1}{c|}{35.45 / 0.9785} & \multicolumn{1}{c|}{29.62 / 0.9302} & \multicolumn{1}{c}{24.16 / 0.8457}        \\
SepConv-ours-e2e              & \multicolumn{1}{c}{\Checkmark}  & \multicolumn{1}{c|}{\XSolidBrush}     & \multicolumn{1}{c|}{33.96 / 0.9650}     & \multicolumn{1}{c|}{26.83 / 0.8639}  & \multicolumn{1}{c|}{40.41 / 0.9911} & \multicolumn{1}{c|}{35.71 / 0.9794} & \multicolumn{1}{c|}{29.80 / 0.9313} & \multicolumn{1}{c}{24.26 / 0.8479} \\
EDSC~\cite{cheng2021multiple}          & \multicolumn{1}{c}{\XSolidBrush} & \multicolumn{1}{c|}{\XSolidBrush}     & \multicolumn{1}{c|}{34.55 / 0.9677}     & \multicolumn{1}{c|}{26.83 / 0.8578}  & \multicolumn{1}{c|}{40.66 / 0.9915} & \multicolumn{1}{c|}{35.77 / 0.9795} & \multicolumn{1}{c|}{29.75 / 0.9301} & \multicolumn{1}{c}{24.12 / 0.8420}        \\
EDSC-ours-e2e                 & \multicolumn{1}{c}{\Checkmark}  & \multicolumn{1}{c|}{\XSolidBrush}     & \multicolumn{1}{c|}{34.73 / 0.9685}     & \multicolumn{1}{c|}{26.96 / 0.8600}  & \multicolumn{1}{c|}{40.88 / 0.9917} & \multicolumn{1}{c|}{35.98 / 0.9803} & \multicolumn{1}{c|}{29.85 / 0.9313} & \multicolumn{1}{c}{24.19 / 0.8436} \\ 
\midrule
RIFE~\cite{huang2022real}              & \multicolumn{1}{c}{\XSolidBrush} & \multicolumn{1}{c|}{\XSolidBrush}     & \multicolumn{1}{c|}{35.28 / 0.9704}     & \multicolumn{1}{c|}{27.61 / 0.8760}  & \multicolumn{1}{c|}{40.74 / 0.9916}  & \multicolumn{1}{c|}{36.18 / 0.9808}                   & \multicolumn{1}{c|}{30.30 / 0.9368}  & \multicolumn{1}{c}{24.62 / 0.8531}        \\
RIFE-ours-e2e                 & \multicolumn{1}{c}{\Checkmark}  & \multicolumn{1}{c|}{\XSolidBrush}     & \multicolumn{1}{c|}{35.57 / 0.9717}     & \multicolumn{1}{c|}{27.81 / 0.8798}  & \multicolumn{1}{c|}{40.95 / 0.9918} & \multicolumn{1}{c|}{36.58 / 0.9816} & \multicolumn{1}{c|}{30.49 / 0.9386} & \multicolumn{1}{c}{24.71 / 0.8549} \\
RIFE-ours-e2e++               & \multicolumn{1}{c}{\Checkmark}  & \multicolumn{1}{c|}{\XSolidBrush}     & \multicolumn{1}{c|}{35.93 / 0.9733}     & \multicolumn{1}{c|}{28.10 / 0.8850}  & \multicolumn{1}{c|}{41.20 / 0.9924} & \multicolumn{1}{c|}{36.94 / 0.9835} & \multicolumn{1}{c|}{30.83 / 0.9430} & \multicolumn{1}{c}{24.87 / 0.8589} \\
RIFE-ours-plugin              & \multicolumn{1}{c}{\XSolidBrush}   & \multicolumn{1}{c|}{\Checkmark}     & \multicolumn{1}{c|}{35.56 / 0.9714}     & \multicolumn{1}{c|}{27.76 / 0.8771}  & \multicolumn{1}{c|}{40.99 / 0.9918} & \multicolumn{1}{c|}{36.55 / 0.9825} & \multicolumn{1}{c|}{30.48 / 0.9387} & \multicolumn{1}{c}{24.64 / 0.8533} \\
\midrule
IFRNet~\cite{kong2022ifrnet}           & \multicolumn{1}{c}{\XSolidBrush} & \multicolumn{1}{c|}{\XSolidBrush}     & \multicolumn{1}{c|}{35.86 / 0.9729}     & \multicolumn{1}{c|}{28.03 / 0.8851}  & \multicolumn{1}{c|}{40.91 / 0.9918}  & \multicolumn{1}{c|}{36.58 / 0.9816} & \multicolumn{1}{c|}{30.75 / 0.9403} & \multicolumn{1}{c}{24.85 / 0.8590}          \\
IFRNet-ours-e2e               & \multicolumn{1}{c}{\Checkmark}  & \multicolumn{1}{c|}{\XSolidBrush}     & \multicolumn{1}{c|}{36.38 / 0.9753}     & \multicolumn{1}{c|}{28.45 / 0.8936}  & \multicolumn{1}{c|}{41.21 / 0.9921} & \multicolumn{1}{c|}{37.03 / 0.9832} & \multicolumn{1}{c|}{31.10 / 0.9440} & \multicolumn{1}{c}{25.03 / 0.8634} \\
IFRNet-ours-e2e++             & \multicolumn{1}{c}{\Checkmark}  & \multicolumn{1}{c|}{\XSolidBrush}     & \multicolumn{1}{c|}{36.68 / 0.9760}     & \multicolumn{1}{c|}{28.78 / 0.8995}  & \multicolumn{1}{c|}{41.48 / 0.9923} & \multicolumn{1}{c|}{37.57 / 0.9850} & \multicolumn{1}{c|}{31.45 / 0.9482} & \multicolumn{1}{c}{25.22 / 0.8694} \\
IFRNet-ours-plugin            & \multicolumn{1}{c}{\XSolidBrush}   & \multicolumn{1}{c|}{\Checkmark}     & \multicolumn{1}{c|}{36.01 / 0.9734}     & \multicolumn{1}{c|}{28.16 / 0.8825}  & \multicolumn{1}{c|}{41.06 / 0.9920} & \multicolumn{1}{c|}{36.92 / 0.9834} & \multicolumn{1}{c|}{30.88 / 0.9404} &  \multicolumn{1}{c}{24.93 / 0.8599} \\
\midrule
UPRNet~\cite{jin2022unified}           & \multicolumn{1}{c}{\XSolidBrush}  & \multicolumn{1}{c|}{\XSolidBrush}     & \multicolumn{1}{c|}{36.07 / 0.9735}     & \multicolumn{1}{c|}{28.38 / 0.8914}  & \multicolumn{1}{c|}{41.01 / 0.9919} & \multicolumn{1}{c|}{36.80 / 0.9819} & \multicolumn{1}{c|}{31.22 / 0.9422} & \multicolumn{1}{c}{25.39 / 0.8648}        \\
UPRNet-ours-e2e              & \multicolumn{1}{c}{\Checkmark}   & \multicolumn{1}{c|}{\XSolidBrush}     & \multicolumn{1}{c|}{36.68 / 0.9758}     & \multicolumn{1}{c|}{\textcolor{blue}{\underline{28.84}} / \textcolor{blue}{\underline{0.8997}}}  & \multicolumn{1}{c|}{41.31 / \textcolor{blue}{\underline{0.9923}}} & \multicolumn{1}{c|}{37.24 / 0.9836} & \multicolumn{1}{c|}{\textcolor{blue}{\underline{31.66}} / 0.9464} & \multicolumn{1}{c}{25.64 / 0.8699} \\
UPRNet-ours-e2e++          & \multicolumn{1}{c}{\Checkmark} & \multicolumn{1}{c|}{\XSolidBrush}     & \multicolumn{1}{c|}{\textcolor{red}{\textbf{36.90}} / \textcolor{blue}{\underline{0.9768}}}     & \multicolumn{1}{c|}{\textcolor{red}{\textbf{29.15}} / \textcolor{red}{\textbf{0.9062}}}  & \multicolumn{1}{c|}{\textcolor{red}{\textbf{41.48}} / \textcolor{red}{\textbf{0.9925}}} & \multicolumn{1}{c|}{\textcolor{red}{\textbf{37.66}} / \textcolor{red}{\textbf{0.9855}}} & \multicolumn{1}{c|}{\textcolor{red}{\textbf{32.00}} / \textcolor{red}{\textbf{0.9519}}} & \multicolumn{1}{c}{\textcolor{red}{\textbf{25.99}} / \textcolor{red}{\textbf{0.8798}}} \\
UPRNet-ours-plugin           & \multicolumn{1}{c}{\XSolidBrush}    & \multicolumn{1}{c|}{\Checkmark}     & \multicolumn{1}{c|}{36.44 / 0.9751}     & \multicolumn{1}{c|}{28.69 / 0.8945}  & \multicolumn{1}{c|}{\textcolor{blue}{\underline{41.32}} / \textcolor{blue}{\underline{0.9923}}} & \multicolumn{1}{c|}{\textcolor{blue}{\underline{37.38}} / 0.9843} & \multicolumn{1}{c|}{31.64 / 0.9448} & \multicolumn{1}{c}{\textcolor{blue}{\underline{25.69}} / 0.8705} \\ 
\midrule
VFIformer~\cite{lu2022vfiformer}       & \multicolumn{1}{c}{\XSolidBrush} & \multicolumn{1}{c|}{\XSolidBrush}     & \multicolumn{1}{c|}{36.14 / 0.9738}     & \multicolumn{1}{c|}{28.33 / 0.8898}  & \multicolumn{1}{c|}{40.93 / 0.9918} & \multicolumn{1}{c|}{36.53 / 0.9815} & \multicolumn{1}{c|}{30.52 / 0.9392} & \multicolumn{1}{c}{24.92 / 0.8580}        \\
EMA-VFI~\cite{Zhang2023ExtractingMA}   & \multicolumn{1}{c}{\XSolidBrush} & \multicolumn{1}{c|}{\XSolidBrush}     & \multicolumn{1}{c|}{36.23 / 0.9740}     & \multicolumn{1}{c|}{28.07 / 0.8826}  & \multicolumn{1}{c|}{41.04 / 0.9921} & \multicolumn{1}{c|}{36.73 / 0.9821} & \multicolumn{1}{c|}{30.88 / 0.9400} & \multicolumn{1}{c}{24.92 / 0.8580}        \\
FLAVR~\cite{kalluri2023flavr}          & \multicolumn{1}{c}{\XSolidBrush} & \multicolumn{1}{c|}{\XSolidBrush}     & \multicolumn{1}{c|}{36.22 / 0.9746}     & \multicolumn{1}{c|}{27.97 / 0.8806}  & \multicolumn{1}{c|}{41.09 / 0.9918} & \multicolumn{1}{c|}{36.85 / 0.9830} & \multicolumn{1}{c|}{31.10 / 0.9456} & \multicolumn{1}{c}{25.23 / 0.8676}        \\
VFIT-S~\cite{shi2022vfit}              & \multicolumn{1}{c}{\XSolidBrush} & \multicolumn{1}{c|}{\XSolidBrush}     & \multicolumn{1}{c|}{36.42 / 0.9760}     & \multicolumn{1}{c|}{28.46 / 0.8926}  & \multicolumn{1}{c|}{41.15 / 0.9920} & \multicolumn{1}{c|}{37.07 / \textcolor{blue}{\underline{0.9845}}} & \multicolumn{1}{c|}{31.39 / \textcolor{blue}{\underline{0.9501}}} & \multicolumn{1}{c}{25.52 / 0.8717}        \\
VFIT-B~\cite{shi2022vfit}              & \multicolumn{1}{c}{\XSolidBrush} & \multicolumn{1}{c|}{\XSolidBrush}     & \multicolumn{1}{c|}{\textcolor{blue}{\underline{36.89}} / \textcolor{red}{\textbf{0.9775}}}     & \multicolumn{1}{c|}{28.60 / 0.8945}  & \multicolumn{1}{c|}{41.24 / 0.9921} & \multicolumn{1}{c|}{37.06 / 0.9839} & \multicolumn{1}{c|}{31.39 / \textcolor{blue}{\underline{0.9501}}} & \multicolumn{1}{c}{25.61 / \textcolor{blue}{\underline{0.8731}}}        \\ 
\bottomrule

\end{tabular}
\end{center}
\vspace{-0.2cm}
\caption{\textbf{Quantitative (PSNR/SSIM) comparison}. We compare our boosted models to representative state-of-the-art methods on Vimeo90K~\cite{xue2019toflow}, DAVIS~\cite{perazzi2016benchmark} and SNU-FILM~\cite{choi2020channel} benchmarks. Both of the optimisation approaches exhibit a substantial improvement in performance. \textbf{Note that} FLAVR~\cite{kalluri2023flavr} and VFIT~\cite{shi2022vfit} take \textbf{multiple frames} as input, but our boosted models can still outperform them. \textcolor{red}{\textbf{RED}}: best performance, \textcolor{blue}{\underline{BLUE}}: second best performance.
}
\label{tab:results}
\vspace{-0.4cm}
\end{table}

As shown in Table~\ref{tab:results}, we can draw the following three observations: 
(i) comparing with the off-the-shelf VFI models, our proposed {\em cycle-consistency adaptation} strategy with end-to-end finetuning can always bring significant PSNR performance gain on all benchmarks, that confirms the universality of our approach;
(ii) the end-to-end adapted IFRNet-ours-e2e and UPRNet-ours-e2e have exhibited comparable performance to state-of-the-art methods, such as EMA-VFI~\cite{Zhang2023ExtractingMA} and {\em etc}. And notably, UPRNet-ours-e2e++ with further adaptation improves \textbf{0.83dB} (36.90dB vs 36.07dB) on Vimeo90K testset and has consistently exhibited a performance gain of over \textbf{0.47dB} on other benchmarks, even outperforming the methods that take multiple frames as input, showing the effectiveness of test-time motion adaptation for unleashing the potential of pre-trained two-frame VFI models; (iii) models with the proposed plug-in {\em adapter} module have exhibited similar performance improvement to end-to-end finetuning, simultaneously incurring efficiency and efficacy.

\begin{figure}[!htb]
  \centering
  \begin{minipage}{\textwidth}
  \centering
  \includegraphics[width=\textwidth]{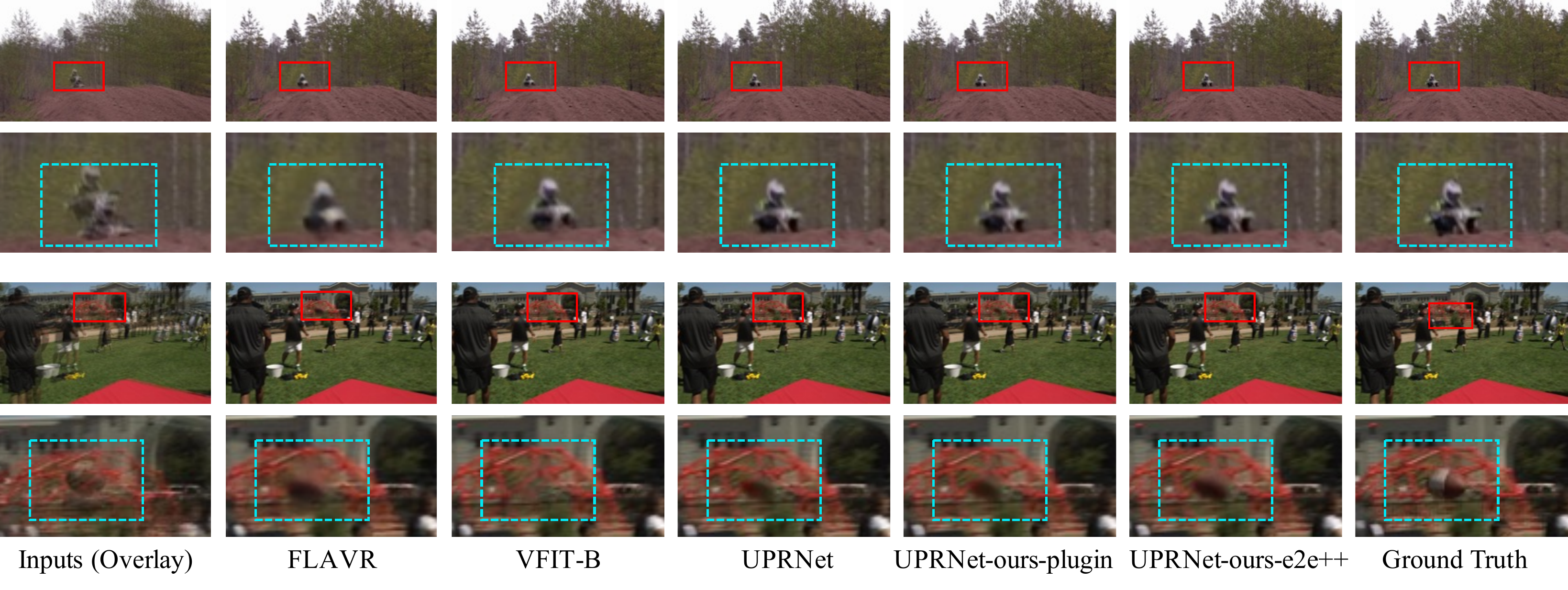} \\
  \vspace{-0.2cm}
  \small
  \caption*{(a) Qualitative Comparison on Vimeo90K~\cite{xue2019toflow}}
  \vspace{0.1cm}
\end{minipage}
  \begin{minipage}{\textwidth}
  \centering
  \includegraphics[width=\textwidth]{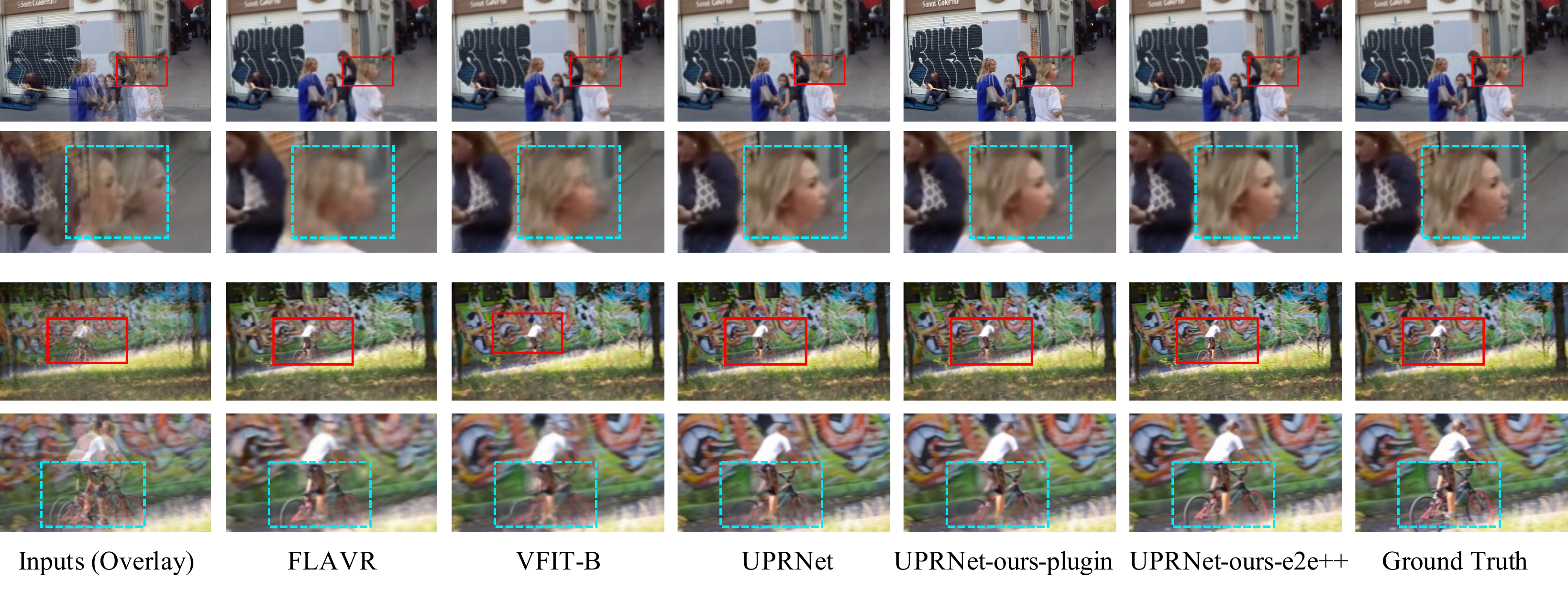} \\
  \vspace{-0.2cm}
  \small 
  \caption*{(b) Qualitative Comparison on SNU-FILM~\cite{choi2020channel} and DAVIS~\cite{perazzi2016benchmark}}
\end{minipage}
    \vspace{-0.1cm}
  \caption{\textbf{Qualitative comparison against the state-of-the-art VFI algorithms.} We show visualization on Vimeo90K~\cite{xue2019toflow}, SNU-FILM~\cite{choi2020channel} and DAVIS~\cite{perazzi2016benchmark} benchmarks for comparison. The patches for careful comparison are marked with \textcolor{red}{red} in the original images. Our boosted models can generate higher-quality results with clearer structures and fewer distortions.}
 \label{fig:results}
 \vspace{-0.4cm}
\end{figure}

\vspace{3pt}
\noindent {\bf Qualitative Results.}  
We demonstrate the qualitative results in Figure~\ref{fig:results}, with the following observations: (i) comparing with existing state-of-the-art methods, the images generated by the models boosted via end-to-end adaptation and plug-in adapter present more details and have higher fidelity; (ii) our test-time optimised models generate less motion blur in the synthesized intermediate frame, indicating that the model has better adapted to the special motion characteristics in each scenario, thus improving the quality of synthesized frames.

\subsection{Ablation Studies}
\label{sec:abl}

In this section, we have conducted thorough ablation studies to quantitatively and qualitatively demonstrate the effectiveness of the proposed {\em cycle-consistency adaptation} strategy and {\em adapter} module from the perspectives of stability and efficiency. 

\vspace{2.5pt}
\noindent{\bf Adaptation Approach.} 
In addition to the {\em cycle-consistency adaptation} proposed by us, 
we here consider a baseline approach for test-time adaptation, 
which is to directly optimise the distance between $\mathcal{I}_3$ and $\hat{\mathcal{I}}_3$ synthesized with $\mathcal{I}_1$ and $\mathcal{I}_5$ as input, denoted as na\"ive optimisation. It is noteworthy that the inter-frame temporal distance during such adaptation is larger than that of test scenario.
As shown in Table \ref{tab:table2}, we compare the two adaption strategies on five VFI methods, and have the following observations: 
(i) under the same adaptation steps, the performance gain of our proposed cycle-consistency adaptation is significantly higher than that of na\"ive adaptation;
(ii) as the steps of adaptation increase, 
na\"ive adaptation may lead to a drop in performance improvement and even result in inferior performance compared to the original pre-trained models, 
whereas cycle-consistency adaptation can steadily boost VFI models, 
as it fully utilizes the inter-frame consistency to learn motion characteristics within the test video sequence. 

\begin{table}[t]
\footnotesize
\tabcolsep=0.1cm
\centering
\begin{tabular}{ccccccc}
\toprule
Strategies & \multicolumn{1}{c|}{\#Adaptations} & \multicolumn{1}{c|}{SepConv~\cite{niklaus2017sepconv}} & \multicolumn{1}{c|}{EDSC~\cite{cheng2021multiple}} & \multicolumn{1}{c|}{RIFE~\cite{huang2022real}} & \multicolumn{1}{c|}{IFRNet~\cite{kong2022ifrnet}} & UPRNet~\cite{jin2022unified} \\ \midrule
Original                & \multicolumn{1}{c|}{0} & \multicolumn{1}{c|}{33.72 / 0.9639} & \multicolumn{1}{c|}{34.55 / 0.9677} & \multicolumn{1}{c|}{35.28 / 0.9704} & \multicolumn{1}{c|}{35.86 / 0.9729} & 36.07 / 0.9735 \\ 
\midrule
\multirow{4}{*}{Na\"ive}  & \multicolumn{1}{c|}{5} & \multicolumn{1}{c|}{33.77 / 0.9641} & \multicolumn{1}{c|}{34.62 / 0.9679} & \multicolumn{1}{c|}{35.36 / 0.9708} & \multicolumn{1}{c|}{35.95 / 0.9734} & 36.23 / 0.9744 \\
                        & \multicolumn{1}{c|}{10} & \multicolumn{1}{c|}{33.83 / 0.9644} & \multicolumn{1}{c|}{34.69 / 0.9683} & \multicolumn{1}{c|}{35.45 / 0.9713} & \multicolumn{1}{c|}{35.81 / 0.9731} & 36.16 / 0.9747 \\ 
                        & \multicolumn{1}{c|}{20} & \multicolumn{1}{c|}{33.91 / 0.9647} & \multicolumn{1}{c|}{34.80 / 0.9687} & \multicolumn{1}{c|}{35.45 / 0.9715} & \multicolumn{1}{c|}{35.03 / 0.9685} & 35.79 / 0.9737 \\ 
                        & \multicolumn{1}{c|}{30} & \multicolumn{1}{c|}{33.95 / 0.9648} & \multicolumn{1}{c|}{34.85 / 0.9688} & \multicolumn{1}{c|}{35.33 / 0.9710} & \multicolumn{1}{c|}{34.09 / 0.9615} & 35.51 / 0.9721 \\ 
 \midrule
\multirow{4}{*}{Cycle}  & \multicolumn{1}{c|}{5} & \multicolumn{1}{c|}{33.83 / 0.9644} & \multicolumn{1}{c|}{34.63 / 0.9680} & \multicolumn{1}{c|}{35.41 / 0.9710} & \multicolumn{1}{c|}{36.14 / 0.9741} & 36.49 / 0.9750 \\
                        & \multicolumn{1}{c|}{10} & \multicolumn{1}{c|}{33.96 / 0.9650} & \multicolumn{1}{c|}{34.73 / 0.9685} & \multicolumn{1}{c|}{35.57 / 0.9717} & \multicolumn{1}{c|}{36.38 / 0.9753} & 36.68 / 0.9758 \\ 
                        & \multicolumn{1}{c|}{20} & \multicolumn{1}{c|}{34.17 / 0.9659} & \multicolumn{1}{c|}{34.94 / 0.9693} & \multicolumn{1}{c|}{35.80 / 0.9728} & \multicolumn{1}{c|}{36.60 / 0.9759} & 36.84 / 0.9766 \\ 
                        & \multicolumn{1}{c|}{30} & \multicolumn{1}{c|}{\textbf{34.29} / \textbf{0.9662}} & \multicolumn{1}{c|}{\textbf{35.06} / \textbf{0.9699}} & \multicolumn{1}{c|}{\textbf{35.93} / \textbf{0.9733}} & \multicolumn{1}{c|}{\textbf{36.68} / \textbf{0.9760}} & \textbf{36.90} / \textbf{0.9768} \\ 
\bottomrule
\end{tabular}
\vspace{5pt}

\caption{\textbf{Quantitative (PSNR/SSIM) comparison of adaptation strategies.} The experiments on Vimeo90K~\cite{xue2019toflow} dataset have shown that cycle-consistency adaptation steadily boosts VFI models by fully leveraging the inter-frame consistency to learn motion characteristics within the test sequence.
}
\label{tab:table2}
\vspace{-0.4cm}
\end{table}





\vspace{2.5pt}
\noindent{\bf Adaptation Cost.} As mentioned in Sec.~\ref{sec:3.3}, the proposed plug-in adapter is designed to improve the efficiency of test-time motion adaptation. 
Here, we conduct end-to-end and plug-in adapter finetuning on three VFI models, 
and compare the number of parameters to be optimised and the time required for each step of adaptation. The results in Table~\ref{table3} have illustrated that with the support of our proposed plug-in adapter, we can achieve a 2 times acceleration with less than 4\% parameters to be optimised, while maintaining inference efficiency and similar quantitative performance improvement comparing to end-to-end finetuning. This confirms the efficiency and feasibility of our proposed plug-in adapter.
\begin{table}[!htb]
\footnotesize
\begin{center}
\tabcolsep=0.15cm
\begin{tabular}{lccccccc}


\toprule
\multicolumn{1}{c}{\multirow{2}{*}{Methods}} & \#Finetuning & \multicolumn{3}{c}{Adaptation Time (ms)} & \multicolumn{3}{c}{Inference Time (ms)} \\ \cline{3-8} 
 & Parameters & Vimeo90K & DAVIS & SNU-FILM & Vimeo90K & DAVIS & SNU-FILM \\
\midrule
\multicolumn{1}{l|}{RIFE-ours-e2e}       & \multicolumn{1}{c|}{10.21M} & \multicolumn{1}{c}{145.6} & \multicolumn{1}{c}{162.7} & \multicolumn{1}{c|}{260.8} & \multicolumn{1}{c}{10.94} & \multicolumn{1}{c}{12.74} & \multicolumn{1}{c}{23.61} \\
\multicolumn{1}{l|}{RIFE-ours-plugin}     & \multicolumn{1}{c|}{0.087M} & \multicolumn{1}{c}{83.13} & \multicolumn{1}{c}{86.84} & \multicolumn{1}{c|}{125.4} & \multicolumn{1}{c}{11.79} & \multicolumn{1}{c}{14.67} & \multicolumn{1}{c}{24.79} \\
\midrule
\multicolumn{1}{l|}{IFRNet-ours-e2e}     & \multicolumn{1}{c|}{18.79M} & \multicolumn{1}{c}{107.7} & \multicolumn{1}{c}{196.2} & \multicolumn{1}{c|}{403.3} & \multicolumn{1}{c}{18.61} & \multicolumn{1}{c}{25.94} & \multicolumn{1}{c}{55.54} \\
\multicolumn{1}{l|}{IFRNet-ours-plugin}   & \multicolumn{1}{c|}{0.676M} & \multicolumn{1}{c}{39.08} & \multicolumn{1}{c}{73.79} & \multicolumn{1}{c|}{158.1} & \multicolumn{1}{c}{19.11} & \multicolumn{1}{c}{29.32} & \multicolumn{1}{c}{61.58} \\
\midrule
\multicolumn{1}{l|}{UPRNet-ours-e2e}     & \multicolumn{1}{c|}{6.260M} & \multicolumn{1}{c}{285.5} & \multicolumn{1}{c}{507.0} & \multicolumn{1}{c|}{1487.8} & \multicolumn{1}{c}{28.33} & \multicolumn{1}{c}{49.90} & \multicolumn{1}{c}{90.85} \\
\multicolumn{1}{l|}{UPRNet-ours-plugin}   & \multicolumn{1}{c|}{0.009M} & \multicolumn{1}{c}{162.0} & \multicolumn{1}{c}{237.6} & \multicolumn{1}{c|}{872.7} & \multicolumn{1}{c}{29.20} & \multicolumn{1}{c}{50.72} & \multicolumn{1}{c}{92.60} \\
\bottomrule

\end{tabular}

\end{center}
\vspace{-0.2cm}
\caption{\textbf{Ablation Study on end-to-end and plug-in adapter adaptation.} Models boosted by our proposed plug-in adapter require minimal finetuning parameters for adaptation, resulting in a 2 times improvement in efficiency while maintaining comparable inference efficiency and performance.}
\label{table3}
\vspace{-0.4cm}
\end{table}

\begin{figure}[t]
\footnotesize
  \centering  
    \includegraphics[width=.99\textwidth]{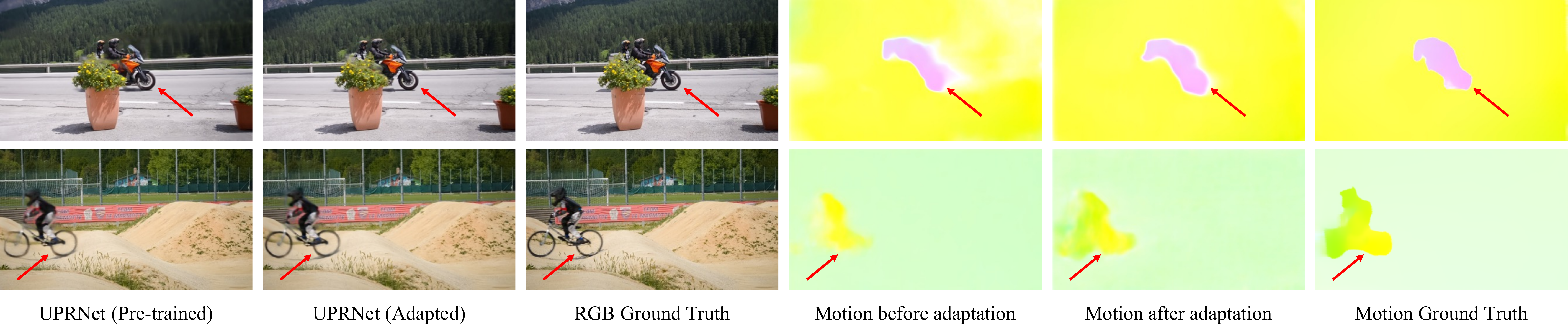}
     \vspace{-0.1cm}
  \caption{\textbf{Motion field visualization.} The VFI model boosted by our proposed motion adaptation can estimate more precise motion fields, thereby producing synthesized frames with higher quality.
  }
 \label{fig:flow_adaptation}
 \vspace{-0.4cm}
\end{figure}

\vspace{2.5pt}
\noindent{\bf Motion Field Visualization.} As stated in Sec.~\ref{sec:3.2}, our proposed cycle-consistency adaptation strategy enables VFI models to fully use inter-frame consistency, and thus acquire motion patterns that are more suitable for testing scenarios.
To qualitatively verify this idea, we visualize the motion fields estimated by the VFI model before and after adaptation.
Specifically, we compute the optical flow between the target frame and the reference frame by RAFT~\cite{teed2020raft} as motion ground truth, 
and compare the motion fields estimated by UPRNet~\cite{jin2022unified} before and after adaptation.
As shown in Figure~\ref{fig:flow_adaptation},  the model boosted by our proposed motion adaptation can output \textbf{smoother motion fields} and \textbf{more precise motion edges}, leading to steady improvement in the quality of synthesized images.

\section{Conclusion}

In this paper, we present optimisation-based video frame interpolation to tackle the generalisation challenge of VFI models and boost their performance at inference time. 
To this end, a test-time motion adaptation strategy that is suitable for video frame interpolation has been introduced, namely {\em cycle-consistency adaptation}. 
In order to address the efficiency drawback of motion adaptation, 
we further propose a lightweight yet effective plug-in {\em adapter} module which can be injected into the motion estimation module of existing pre-trained VFI models to refine the estimated motion flow, thus synthesizing higher-quality intermediate frames. 
Extensive experiments on various models and benchmarks have demonstrated the effectiveness of the proposed cycle-consistency adaptation strategy on VFI task and confirmed that the proposed plug-in adapter module can efficiently and steadily boost the performance of VFI models, even outperforming approaches with extra inputs. 



\section*{Acknowledgments}
This work is supported by National Natural Science Foundation of China (62271308), STCSM (22511105700, 22DZ2229005), 111 plan (BP0719010), and State Key Laboratory of UHD Video and Audio Production and Presentation.

\bibliography{egbib}

\clearpage

\appendix

\noindent{\bf \color{bmvc_blue} \Large Appendix}

\vspace{0.3cm}

In this supplementary document, we start by giving more details on the implementation details of our proposed cycle-consistency motion adaptation strategy and plug-in adapter module in Section~\ref{implementation}. 
Then, we demonstrate the ablation study on the influence of more adaptation steps and how cycle-consistency adaptation steadily works on several video frame interpolation (VFI) models in Section~\ref{ablation}. 
Next, we supplement more qualitative comparison results on several benchmarks in Section~\ref{qualitative}. 
Finally, we illustrate the limitation of our method and our future work in Section~\ref{limitation}.

\section{Implementation Details}
\label{implementation}
\noindent{\textbf{Cycle-consistency adaptation.}} Considering that each sample in common datasets~\cite{xue2019toflow, perazzi2016benchmark, choi2020channel} typically comprises a septuplet, with odd frames as input, denoted as $\{\mathcal{I}_1, \mathcal{I}_3, \mathcal{I}_5, \mathcal{I}_7\}$. 
We divide it into two triplets, {\em i.e.,} $\{\mathcal{I}_1, \mathcal{I}_3, \mathcal{I}_5\}$ and $\{\mathcal{I}_3, \mathcal{I}_5, \mathcal{I}_7\}$ and perform motion adaptation on these two triplets to adapt the model to the motion characteristics of the current video sequence.
To be specific, we take the middle frame of each triplet as the ground truth and calculate loss between the generated frames and the target frames, denoted as $\mathcal{L}_3 = ||\hat{\mathcal{I}}_3 - \mathcal{I}_3||$ and $\mathcal{L}_5 = ||\hat{\mathcal{I}}_5 - \mathcal{I}_5||$. 
Then we take their average $\mathcal{L} = \frac{1}{2}(\mathcal{L}_3 + \mathcal{L}_5)$ to backward gradients for parameters updates, and regard this entire process as one adaptation step. This cycle-consistency adaptation strategy has been employed in end-to-end finetuning as well as plug-in adapter finetuning, and theoretically, it can be extended to longer video sequences.

\vspace{0.2cm}
\noindent{\textbf{Plug-in adapter.}} For flow-based VFI methods~\cite{huang2022real, kong2022ifrnet, jin2022unified}, we can freeze the pre-trained parameters and simply use our proposed lightweight plug-in adapter to adjust the estimated motion flow. Concretely, we reuse the convolutional layers from the pre-trained model's motion estimation module to align the channel dimensions of the extracted visual features with that of motion flow. Then a simple $1 \times 1$ convolution layer is utilized to predict pixel-wise weights $\alpha$ and biases $\beta$ for motion adaptation, without introducing any additional activation layers.

\begin{table}[t]
\scriptsize
\begin{center}
\tabcolsep=0.06cm
\begin{tabular}{lccccccc}
\toprule
\multicolumn{1}{c}{\multirow{2}{*}{Methods}} & \multicolumn{1}{c}{\#Adapt}  & \multirow{2}{*}{Vimeo90K~\cite{xue2019toflow}} & \multirow{2}{*}{DAVIS~\cite{perazzi2016benchmark}} & \multicolumn{4}{c}{SNU-FILM~\cite{choi2020channel}}                                                    \\ 
\cmidrule{5-8}
\multicolumn{1}{c}{}   &  Steps  &                        &                           &                        Easy                  & Medium                & Hard                  & Extreme \\ 
\midrule
\multicolumn{1}{l|}{\multirow{4}{*}{SepConv-ours-e2e}} &   
                            \multicolumn{1}{c|}{0} & \multicolumn{1}{c|}{33.72 / 0.9639} & \multicolumn{1}{c|}{26.65 / 0.8611} & \multicolumn{1}{c|}{40.21 / 0.9909} & \multicolumn{1}{c|}{35.45 / 0.9785} & \multicolumn{1}{c|}{29.62 / 0.9302} & 24.16 / 0.8457 \\
    \multicolumn{1}{l|}{} & \multicolumn{1}{c|}{10} & \multicolumn{1}{c|}{33.96 / 0.9650} & \multicolumn{1}{c|}{26.83 / 0.8639} & \multicolumn{1}{c|}{40.41 / 0.9911} & \multicolumn{1}{c|}{35.71 / 0.9794} & \multicolumn{1}{c|}{29.80 / 0.9313} & 24.26 / 0.8479 \\
    \multicolumn{1}{l|}{} & \multicolumn{1}{c|}{20} & \multicolumn{1}{c|}{34.17 / 0.9659} & \multicolumn{1}{c|}{26.93 / 0.8649} & \multicolumn{1}{c|}{40.56 / 0.9913} & \multicolumn{1}{c|}{35.91 / 0.9800} & \multicolumn{1}{c|}{29.88 / 0.9313} & 24.32 / 0.8484 \\
    \multicolumn{1}{l|}{} & \multicolumn{1}{c|}{30} & \multicolumn{1}{c|}{\textbf{34.29} / \textbf{0.9662}} & \multicolumn{1}{c|}{\textbf{26.98} / \textbf{0.8651}} & \multicolumn{1}{c|}{\textbf{40.66} / \textbf{0.9914}} & \multicolumn{1}{c|}{\textbf{36.04} / \textbf{0.9804}} & \multicolumn{1}{c|}{\textbf{29.93} / \textbf{0.9318}} & \textbf{24.36} / \textbf{0.8491} \\ 
\midrule
\multicolumn{1}{l|}{\multirow{4}{*}{EDSC-ours-e2e}} &   
                            \multicolumn{1}{c|}{0} & \multicolumn{1}{c|}{34.55 / 0.9677} & \multicolumn{1}{c|}{26.83 / 0.8578} & \multicolumn{1}{c|}{40.66 / 0.9915} & \multicolumn{1}{c|}{35.77 / 0.9795} & \multicolumn{1}{c|}{29.75 / 0.9301} & 24.12 / 0.8420 \\
    \multicolumn{1}{l|}{} & \multicolumn{1}{c|}{10} & \multicolumn{1}{c|}{34.73 / 0.9685} & \multicolumn{1}{c|}{26.96 / 0.8600} & \multicolumn{1}{c|}{40.88 / 0.9917} & \multicolumn{1}{c|}{35.98 / 0.9803} & \multicolumn{1}{c|}{29.85 / 0.9313} & 24.19 / 0.8436 \\
    \multicolumn{1}{l|}{} & \multicolumn{1}{c|}{20} & \multicolumn{1}{c|}{34.94 / 0.9693} & \multicolumn{1}{c|}{27.07 / 0.8618} & \multicolumn{1}{c|}{40.98 / 0.9919} & \multicolumn{1}{c|}{36.18 / 0.9811} & \multicolumn{1}{c|}{29.95 / 0.9322} & 24.23 / 0.8445 \\
    \multicolumn{1}{l|}{} & \multicolumn{1}{c|}{30} & \multicolumn{1}{c|}{\textbf{35.06} / \textbf{0.9699}} & \multicolumn{1}{c|}{\textbf{27.14} / \textbf{0.8630}} & \multicolumn{1}{c|}{\textbf{41.09} / \textbf{0.9920}} & \multicolumn{1}{c|}{\textbf{36.33} / \textbf{0.9817}} & \multicolumn{1}{c|}{\textbf{30.02} / \textbf{0.9334}} & \textbf{24.28} / \textbf{0.8455} \\ 
\midrule
\multicolumn{1}{l|}{\multirow{4}{*}{RIFE-ours-e2e}} &   
                            \multicolumn{1}{c|}{0} & \multicolumn{1}{c|}{35.28 / 0.9704} & \multicolumn{1}{c|}{27.61 / 0.8760} & \multicolumn{1}{c|}{40.75 / 0.9916} & \multicolumn{1}{c|}{36.18 / 0.9808} & \multicolumn{1}{c|}{30.30 / 0.9368} & 24.62 / 0.8531 \\
    \multicolumn{1}{l|}{} & \multicolumn{1}{c|}{10} & \multicolumn{1}{c|}{35.57 / 0.9717} & \multicolumn{1}{c|}{27.81 / 0.8798} & \multicolumn{1}{c|}{40.95 / 0.9918} & \multicolumn{1}{c|}{36.42 / 0.9815} & \multicolumn{1}{c|}{30.49 / 0.9386} & 24.71 / 0.8549 \\
    \multicolumn{1}{l|}{} & \multicolumn{1}{c|}{20} & \multicolumn{1}{c|}{35.80 / 0.9728} & \multicolumn{1}{c|}{27.99 / 0.8830} & \multicolumn{1}{c|}{41.10 / 0.9920} & \multicolumn{1}{c|}{36.71 / 0.9827} & \multicolumn{1}{c|}{30.67 / 0.9408} & 24.79 / 0.8570 \\
    \multicolumn{1}{l|}{} & \multicolumn{1}{c|}{30} & \multicolumn{1}{c|}{\textbf{35.93} / \textbf{0.9733}} & \multicolumn{1}{c|}{\textbf{28.10} / \textbf{0.8850}} & \multicolumn{1}{c|}{\textbf{41.20} / \textbf{0.9924}} & \multicolumn{1}{c|}{\textbf{36.94} / \textbf{0.9835}} & \multicolumn{1}{c|}{\textbf{30.83} / \textbf{0.9430}} & \textbf{24.87} / \textbf{0.8589} \\ 
\midrule
\multicolumn{1}{l|}{\multirow{4}{*}{RIFE-ours-plugin}} & 
                            \multicolumn{1}{c|}{0} & \multicolumn{1}{c|}{35.33 / 0.9706} & \multicolumn{1}{c|}{27.64 / 0.8765} & \multicolumn{1}{c|}{40.66 / 0.9915} & \multicolumn{1}{c|}{36.12 / 0.9807} & \multicolumn{1}{c|}{30.32 / 0.9371} & 24.67 / 0.8539 \\
    \multicolumn{1}{l|}{} & \multicolumn{1}{c|}{10} & \multicolumn{1}{c|}{35.56 / 0.9714} & \multicolumn{1}{c|}{27.76 / 0.8871} & \multicolumn{1}{c|}{40.99 / 0.9918} & \multicolumn{1}{c|}{36.55 / 0.9825} & \multicolumn{1}{c|}{30.48 / 0.9387} & 24.64 / 0.8533 \\
    \multicolumn{1}{l|}{} & \multicolumn{1}{c|}{20} & \multicolumn{1}{c|}{35.61 / 0.9719} & \multicolumn{1}{c|}{27.79 / 0.8786} & \multicolumn{1}{c|}{41.01 / 0.9919} & \multicolumn{1}{c|}{36.55 / 0.9824} & \multicolumn{1}{c|}{30.65 / 0.9404} & 24.79 / 0.8555 \\
    \multicolumn{1}{l|}{} & \multicolumn{1}{c|}{30} & \multicolumn{1}{c|}{\textbf{35.71} / \textbf{0.9722}} & \multicolumn{1}{c|}{\textbf{27.88} / \textbf{0.8799}} & \multicolumn{1}{c|}{\textbf{41.12} / \textbf{0.9920}} & \multicolumn{1}{c|}{\textbf{36.77} / \textbf{0.9832}} & \multicolumn{1}{c|}{\textbf{30.74} / \textbf{0.9404}} & \textbf{24.84} / \textbf{0.8590}\\ 
\midrule
\multicolumn{1}{l|}{\multirow{4}{*}{IFRNet-ours-e2e}} & 
                            \multicolumn{1}{c|}{0} & \multicolumn{1}{c|}{35.86 / 0.9729} & \multicolumn{1}{c|}{28.03 / 0.8851} & \multicolumn{1}{c|}{40.91 / 0.9916} & \multicolumn{1}{c|}{36.18 / 0.9808} & \multicolumn{1}{c|}{30.30 / 0.9368} & 24.62 / 0.8531 \\
    \multicolumn{1}{l|}{} & \multicolumn{1}{c|}{10} & \multicolumn{1}{c|}{36.38 / 0.9753} & \multicolumn{1}{c|}{28.45 / 0.8936} & \multicolumn{1}{c|}{41.21 / 0.9921} & \multicolumn{1}{c|}{37.03 / 0.9832} & \multicolumn{1}{c|}{31.10 / 0.9440} & 25.03 / 0.8634 \\
    \multicolumn{1}{l|}{} & \multicolumn{1}{c|}{20} & \multicolumn{1}{c|}{36.60 / 0.9759} & \multicolumn{1}{c|}{28.69 / 0.8979} & \multicolumn{1}{c|}{41.40 / 0.9923} & \multicolumn{1}{c|}{37.36 / 0.9844} & \multicolumn{1}{c|}{31.37 / 0.9476} & 25.18 / 0.8676 \\
    \multicolumn{1}{l|}{} & \multicolumn{1}{c|}{30} & \multicolumn{1}{c|}{\textbf{36.68} / \textbf{0.9760}} & \multicolumn{1}{c|}{\textbf{28.78} / \textbf{0.8995}} & \multicolumn{1}{c|}{\textbf{41.48} / \textbf{0.9923}} & \multicolumn{1}{c|}{\textbf{37.57} / \textbf{0.9850}} & \multicolumn{1}{c|}{\textbf{31.45} / \textbf{0.9482}} & \textbf{25.22} / \textbf{0.8694} \\ 
\midrule
\multicolumn{1}{l|}{\multirow{4}{*}{IFRNet-ours-plugin}} & 
                            \multicolumn{1}{c|}{0} & \multicolumn{1}{c|}{35.86 / 0.9729} & \multicolumn{1}{c|}{28.02 / 0.8850} & \multicolumn{1}{c|}{40.91 / 0.9918} & \multicolumn{1}{c|}{36.58 / 0.9816} & \multicolumn{1}{c|}{30.74 / 0.9404} & 24.84 / 0.8590 \\
    \multicolumn{1}{l|}{} & \multicolumn{1}{c|}{10} & \multicolumn{1}{c|}{36.01 / 0.9734} & \multicolumn{1}{c|}{28.16 / 0.8825} & \multicolumn{1}{c|}{41.06 / 0.9920} & \multicolumn{1}{c|}{36.92 / 0.9834} & \multicolumn{1}{c|}{30.88 / 0.9404} & 24.93 / 0.8599 \\
    \multicolumn{1}{l|}{} & \multicolumn{1}{c|}{20} & \multicolumn{1}{c|}{36.11 / 0.9738} & \multicolumn{1}{c|}{28.26 / 0.8875} & \multicolumn{1}{c|}{41.11 / 0.9921} & \multicolumn{1}{c|}{37.01 / 0.9837} & \multicolumn{1}{c|}{30.95 / 0.9414} & 24.96 / 0.8599 \\
    \multicolumn{1}{l|}{} & \multicolumn{1}{c|}{30} & \multicolumn{1}{c|}{\textbf{36.14} / \textbf{0.9742}} & \multicolumn{1}{c|}{\textbf{28.33} / \textbf{0.8888}} & \multicolumn{1}{c|}{\textbf{41.18} / \textbf{0.9923}} & \multicolumn{1}{c|}{\textbf{37.14} / \textbf{0.9844}} & \multicolumn{1}{c|}{\textbf{31.03} / \textbf{0.9419}} & \textbf{24.97} / \textbf{0.8600} \\ 
\midrule
\multicolumn{1}{l|}{\multirow{4}{*}{UPRNet-ours-e2e}} & 
                            \multicolumn{1}{c|}{0} & \multicolumn{1}{c|}{36.07 / 0.9735} & \multicolumn{1}{c|}{28.38 / 0.8914} & \multicolumn{1}{c|}{41.01 / 0.9919} & \multicolumn{1}{c|}{36.80 / 0.9819} & \multicolumn{1}{c|}{31.22 / 0.9422} & 25.39 / 0.8648 \\
    \multicolumn{1}{l|}{} & \multicolumn{1}{c|}{10} & \multicolumn{1}{c|}{36.68 / 0.9758} & \multicolumn{1}{c|}{28.84 / 0.8997} & \multicolumn{1}{c|}{41.31 / 0.9923} & \multicolumn{1}{c|}{37.24 / 0.9836} & \multicolumn{1}{c|}{31.66 / 0.9464} & 25.64 / 0.8699 \\
    \multicolumn{1}{l|}{} & \multicolumn{1}{c|}{20} & \multicolumn{1}{c|}{36.84 / 0.9766} & \multicolumn{1}{c|}{29.07 / 0.9043} & \multicolumn{1}{c|}{41.42 / 0.9924} & \multicolumn{1}{c|}{37.52 / 0.9849} & \multicolumn{1}{c|}{31.89 / 0.9500} & 25.85 / 0.8755 \\
    \multicolumn{1}{l|}{} & \multicolumn{1}{c|}{30} & \multicolumn{1}{c|}{\textbf{36.90} / \textbf{0.9768}} & \multicolumn{1}{c|}{\textbf{29.15} / \textbf{0.9062}} & \multicolumn{1}{c|}{\textbf{41.48} / \textbf{0.9925}} & \multicolumn{1}{c|}{\textbf{37.66} / \textbf{0.9855}} & \multicolumn{1}{c|}{\textbf{32.00} / \textbf{0.9519}} & \textbf{25.99} / \textbf{0.8798} \\ 
\midrule
\multicolumn{1}{l|}{\multirow{4}{*}{UPRNet-ours-plugin}} & 
                            \multicolumn{1}{c|}{0} & \multicolumn{1}{c|}{36.04 / 0.9734} & \multicolumn{1}{c|}{28.31 / 0.8896} & \multicolumn{1}{c|}{41.00 / 0.9919} & \multicolumn{1}{c|}{36.77 / 0.9818} & \multicolumn{1}{c|}{31.18 / 0.9418} & 25.38 / 0.8645 \\
    \multicolumn{1}{l|}{} & \multicolumn{1}{c|}{10} & \multicolumn{1}{c|}{36.44 / 0.9751} & \multicolumn{1}{c|}{28.69 / 0.8945} & \multicolumn{1}{c|}{41.32 / 0.9923} & \multicolumn{1}{c|}{37.38 / 0.9843} & \multicolumn{1}{c|}{31.64 / 0.9448} & \textbf{25.69} / 0.8705 \\
    \multicolumn{1}{l|}{} & \multicolumn{1}{c|}{20} & \multicolumn{1}{c|}{36.52 / 0.9754} & \multicolumn{1}{c|}{28.78 / 0.8963} & \multicolumn{1}{c|}{41.37 / 0.9923} & \multicolumn{1}{c|}{\textbf{37.59} / 0.9847} & \multicolumn{1}{c|}{31.70 / 0.9461} & 25.68 / 0.8705 \\
    \multicolumn{1}{l|}{} & \multicolumn{1}{c|}{30} & \multicolumn{1}{c|}{\textbf{36.57} / \textbf{0.9756}} & \multicolumn{1}{c|}{\textbf{28.90} / \textbf{0.8989}} & \multicolumn{1}{c|}{\textbf{41.40} / \textbf{0.9924}} & \multicolumn{1}{c|}{37.55 / \textbf{0.9849}} & \multicolumn{1}{c|}{\textbf{31.75} / \textbf{0.9470}} & \textbf{25.69} / \textbf{0.8706} \\ 
\bottomrule
\end{tabular}
\end{center}
\vspace{-0.2cm}
\caption{\textbf{Additional ablation study on adaptation strategies and steps}. 
We inherit pre-trained VFI models~\cite{niklaus2017sepconv, cheng2021multiple, huang2022real, kong2022ifrnet, jin2022unified}, then perform end-to-end (e2e) and plug-in adapter (plugin) adaptation with different adaptation steps, comparing their performances on three commonly used benchmarks.
}
\label{tab:supp}
\vspace{-0.4cm}
\end{table}

\section{Additional Ablation Study}
\label{ablation}
In this section, we present more ablation results for our proposed optimisation-based VFI, including the motion adaptation strategy and the adaptation steps.
Specifically, we conduct test-time motion adaptation experiments on three benchmarks, namely Vimeo90K~\cite{xue2019toflow}, DAVIS~\cite{perazzi2016benchmark}, and SNU-FILM~\cite{choi2020channel} with five VFI models, SepConv~\cite{niklaus2017sepconv}, EDSC~\cite{cheng2021multiple}, RIFE~\cite{huang2022real}, IFRNet~\cite{kong2022ifrnet} and UPRNet~\cite{jin2022unified}, in two manners, end-to-end (e2e) and plug-in adapter (plugin) boosted. We further investigate the impact of different adaptation steps on performance.

\vspace{0.2cm}
\noindent{\textbf{Adaptation Strategy.}} The results in Table~\ref{tab:supp} further demonstrate the conclusions we have drawn in our paper: The performance of pre-trained VFI models can be enhanced via end-to-end motion adaptation during test-time. 
Furthermore, for flow-based methods such as RIFE~\cite{huang2022real}, IFRNet~\cite{kong2022ifrnet} and UPRNet~\cite{jin2022unified}, the proposed plug-in adapter module can effectively boost the performance of pre-trained models. 
Moreover, it achieves comparable performance improvements to that of end-to-end finetuning, with the same number of adaptation steps.

\vspace{0.2cm}
\noindent{\textbf{Adaptation Steps.}} As indicated in Table~\ref{tab:supp}, with adaptation steps increasing, both end-to-end finetuning and plug-in adapter finetuning effectively boost the performance of VFI models, thus confirming the effectiveness of our proposed optimisation-based VFI. 
Furthermore, while cycle-consistency adaptation can steadily boost performance, models exhibit significant improvements within the first $10$ adaptation steps, with performance gains gradually approaching saturation after more steps. Therefore, taking both efficiency and performance into consideration, we have chosen $10$ steps of adaptation as our default setting.

\section{More Qualitative Results}
\label{qualitative}

\begin{figure}[!htbp]
  \centering
  \includegraphics[width=\textwidth]{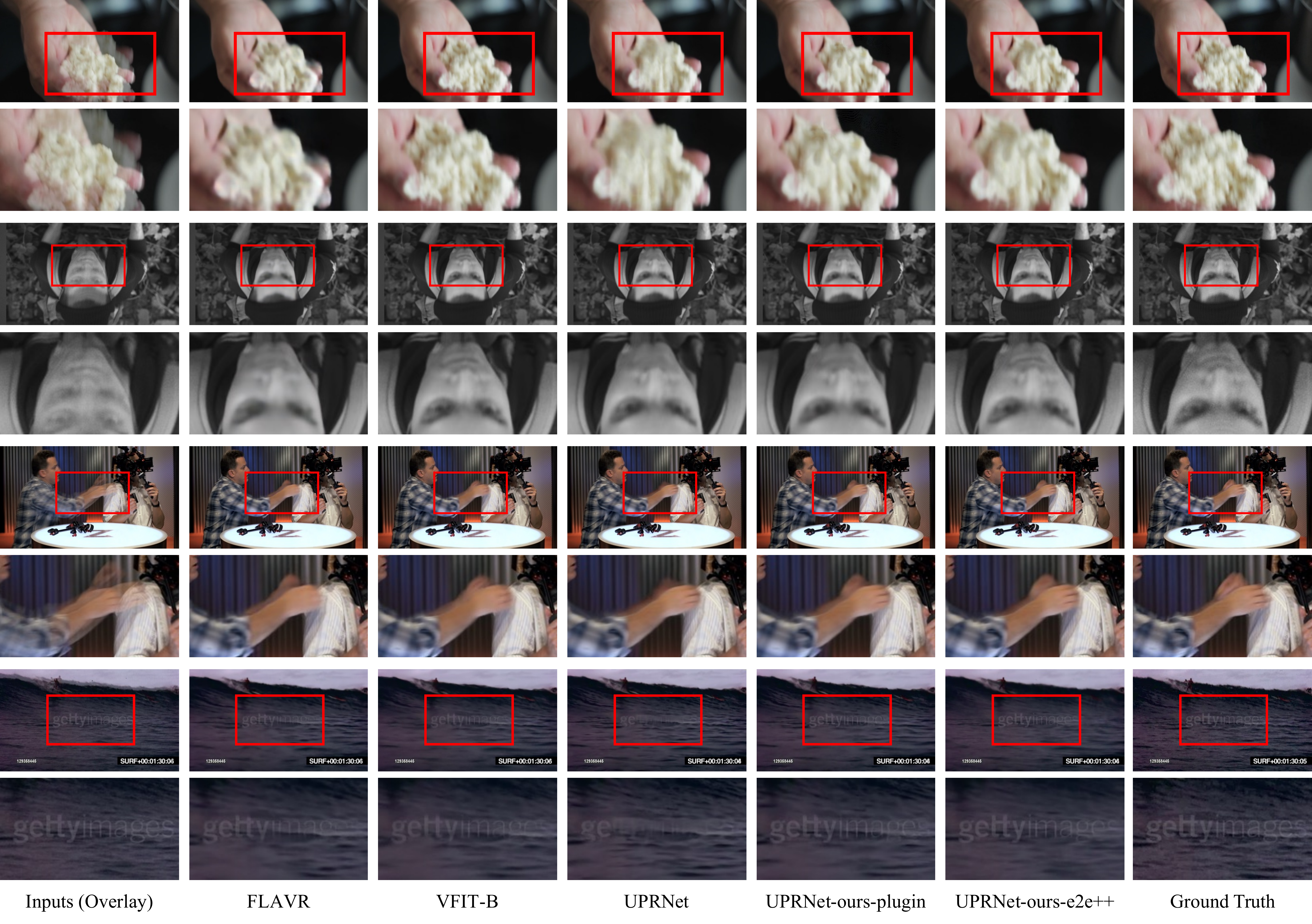} \\
    \vspace{-0.1cm}
  \caption{\textbf{Qualitative comparison against the state-of-the-art VFI algorithms.} We show visualization on Vimeo90K~\cite{xue2019toflow} benchmark for comparison. The patches for careful comparison are marked with \textcolor{red}{red} in the original images. Our boosted models can generate higher-quality results with clearer structures and fewer distortions.}
 \label{fig:supp_figure1}
 \vspace{-0.4cm}
\end{figure}

\begin{figure}[!htbp]
  \centering
  \includegraphics[width=\textwidth]{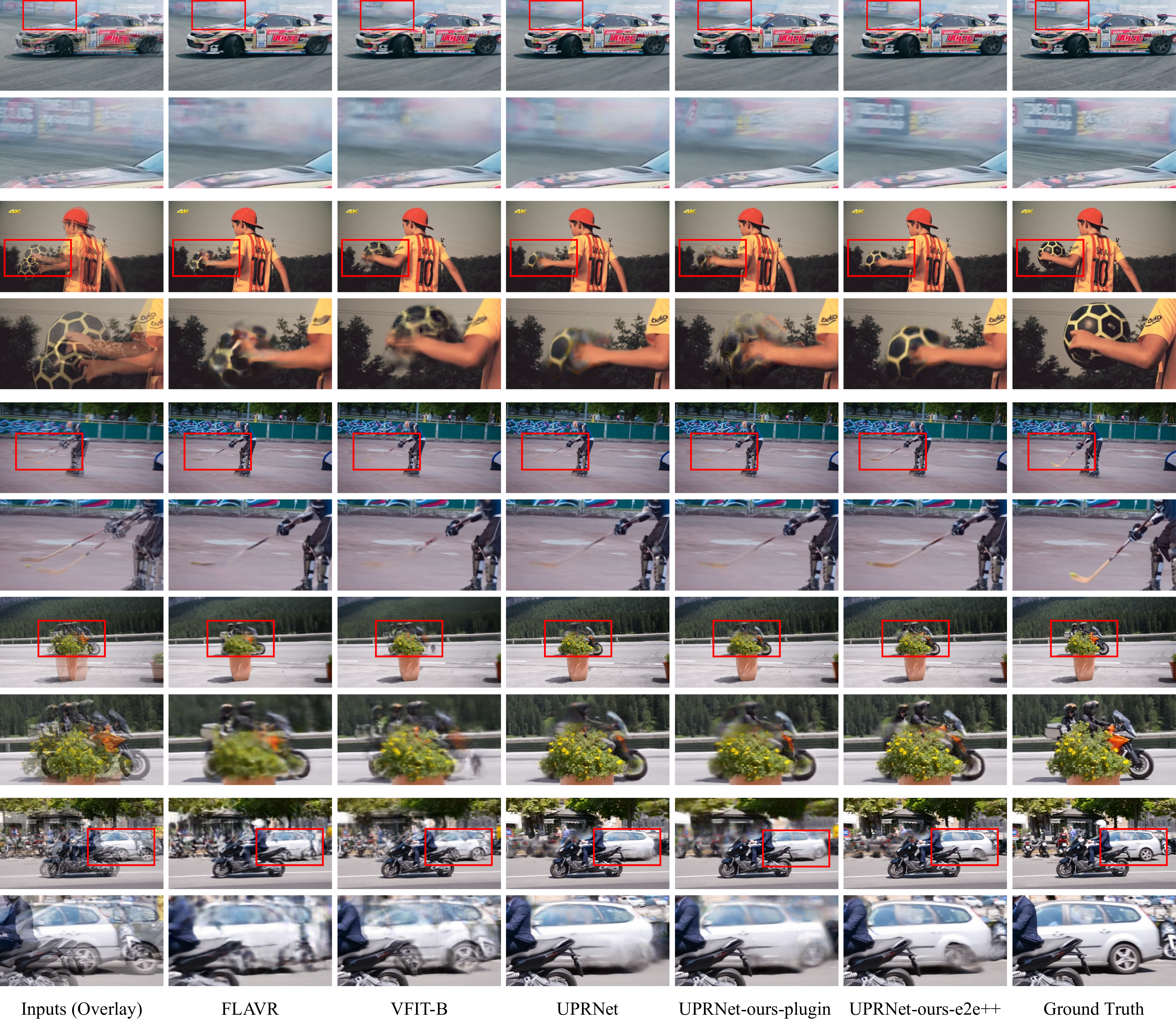} \\
    \vspace{-0.1cm}
  \caption{\textbf{Qualitative comparison against the state-of-the-art VFI algorithms.} We show visualization on DAVIS~\cite{perazzi2016benchmark} benchmark for comparison. The patches for careful comparison are marked with \textcolor{red}{red} in the original images. Our boosted models can generate higher-quality results with clearer structures and fewer distortions.}
 \label{fig:supp_figure2}
 \vspace{-0.4cm}
\end{figure}

\begin{figure}[!htbp]
  \centering
  \includegraphics[width=\textwidth]{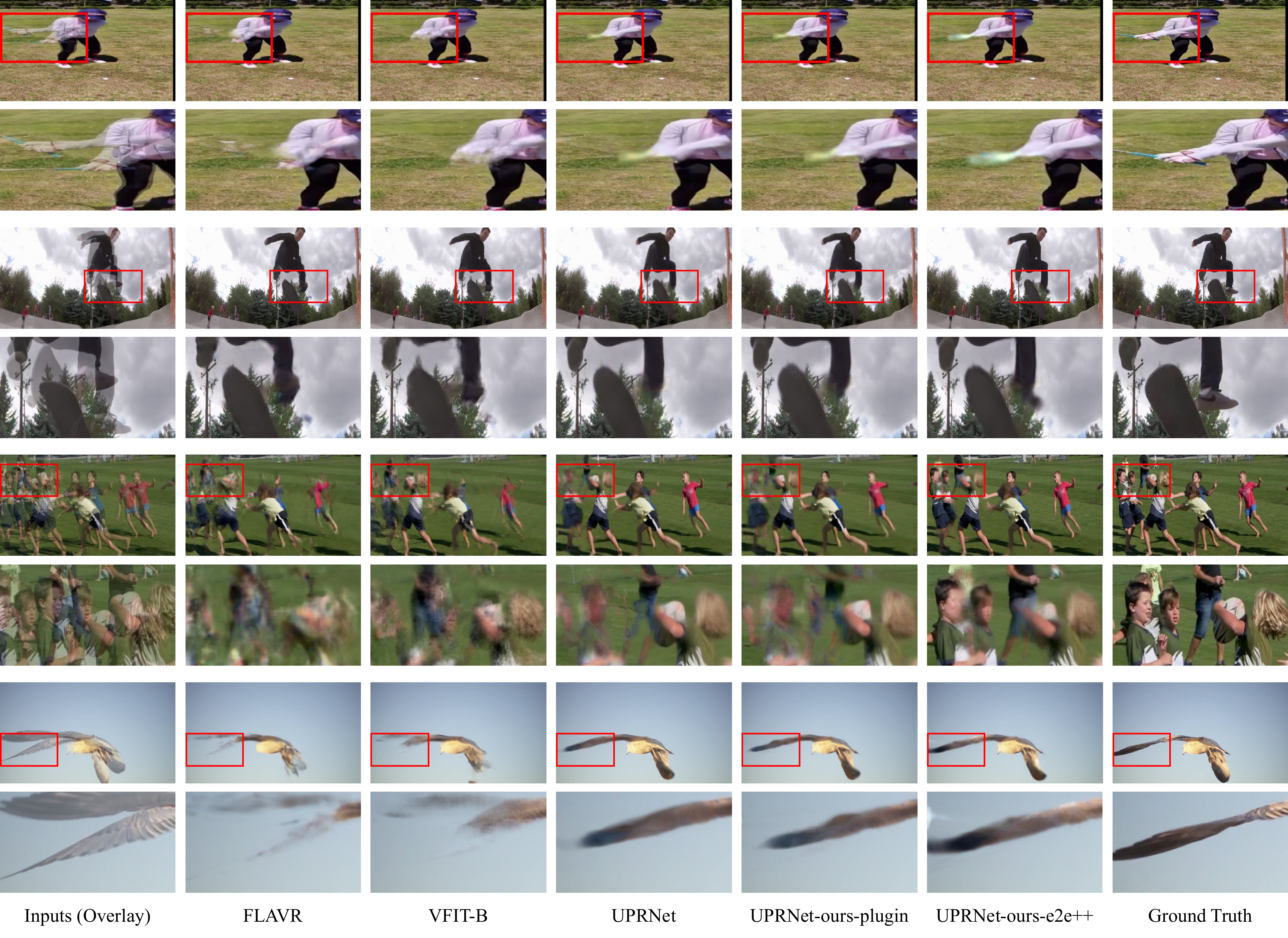} \\
    \vspace{-0.1cm}
  \caption{\textbf{Qualitative comparison against the state-of-the-art VFI algorithms.} We show visualization on SNU-FILM~\cite{choi2020channel} benchmark for comparison. The patches for careful comparison are marked with \textcolor{red}{red} in the original images. Our boosted models can generate higher-quality results with clearer structures and fewer distortions.}
 \label{fig:supp_figure3}
 \vspace{-0.4cm}
\end{figure}

We provide more qualitative results in Figure~\ref{fig:supp_figure1}, Figure~\ref{fig:supp_figure2} and Figure~\ref{fig:supp_figure3}. 
Concretely, we compare UPRNet-ours-plugin (10-step-adaptation) and UPRNet-ours-e2e++ (30-step-adaptation) with other state-of-the-art VFI methods on benchmark datasets with different motion characteristics. 
We can observe that other VFI methods tend to generate images with motion blur or missing object details, while our method can consistently achieve better perceptual quality, {\em i.e.,} maintaining the object shape and textual information, and synthesizing shaper details. 
In summary, our proposed motion adaptation can steadily boost VFI models.

\section{Limitation \& Future Work}
\label{limitation}
In this paper, we have demonstrated the effectiveness of our proposed cycle-consistency motion adaptation and lightweight adapter as a plug-in module for VFI. 
However, some limitations remain: it still suffers from the well-known problem of using test-time adaptation, that is, time consumption is still non-negligible, which poses limitations in increasing adaptation steps to pursue higher performance. 
Besides, due to the limitation of data and computing resources, our motion adaptation strategy is currently restricted to 2-frame VFI models and has not been extended to frame synthesis at arbitrary temporal positions. 
Moreover, in general, our proposed method is beneficial for better handling motions that are regular, such as rotations and large-scale motions, however, it remains challenging for irregular motions, such as sudden camera shaking, or illumination changes, due to the difficulty of inferring priors.
Our future work will investigate more efficient plug-in adapter architecture and further extend it to flexible VFI models with multiple frames as inputs, which has the potential to further boost the performance and generalisation ability of video frame interpolation models.

\end{document}